\documentclass{article}

\usepackage{arxiv}

\usepackage[utf8]{inputenc} 
\usepackage[T1]{fontenc}    
\usepackage{hyperref}       
\usepackage{url}            
\usepackage{booktabs}       
\usepackage{amsfonts}       
\usepackage{nicefrac}       
\usepackage{microtype}      
\usepackage{lipsum}		
\usepackage{graphicx}
\usepackage{doi}

\usepackage[numbers]{natbib}
\usepackage{caption}
\usepackage{algorithm}
\usepackage{algorithmic}
\usepackage{color,soul}
\usepackage{comment}
\usepackage{epstopdf}
\usepackage{float}
\usepackage{amsmath}
\usepackage{bm}
\usepackage{tablefootnote}
\usepackage[normalem]{ulem}
\usepackage{multirow}
\usepackage{amssymb}
\usepackage{graphicx}
\usepackage{booktabs,subcaption,amsfonts,dcolumn}
\usepackage{array}

\definecolor{amber}{rgb}{1.0, 0.49, 0.0}

\newif\ifCommentsAuthors

\CommentsAuthorstrue
\ifCommentsAuthors

\newif\ifshowcomm
\showcommtrue

\definecolor{myred}{rgb}{.8,.0,.0}

\definecolor{myblue}{rgb}{0,0,.8}

\definecolor{mcolor}{rgb}{0,0.5,0.1}

\definecolor{mcolor}{rgb}{0.5,0.2,0.1}
\definecolor{mygreen}{rgb}{.35,0.7,.35}
\definecolor{myblack}{rgb}{0,0,0}
\definecolor{myorange}{rgb}{1,0.5,0}
\newcommand{\ctony}[1]{\ifshowcomm \textcolor{myblack}{#1} \fi}

\else
    
\fi

\title{Energy Disaggregation using Variational Autoencoders}

\author{ 
    \href{https://orcid.org/0000-0002-5837-1475}{\includegraphics[scale=0.06]{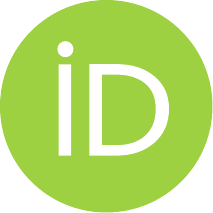}\hspace{1mm}Antoine~Langevin} \\
	Department of Electrical Engineering\\
	\'Ecole de Technologie Sup\'erieure, \\ Universit\'e du Qu\'ebec\\
	Montr\'eal, QC, Canada \\
	\texttt{antoine.langevin.1@etsmtl.net} \\
	\And
	\href{https://orcid.org/0000-0002-0677-415X}{\includegraphics[scale=0.06]{orcid.pdf}\hspace{1mm}Marc-André~Carbonneau} \\
	La Forge Research Laboratory - Ubisoft\\
	Montr\'eal, QC, Canada \\
	\And
	\href{https://orcid.org/0000-0002-5246-7265}{\includegraphics[scale=0.06]{orcid.pdf}\hspace{1mm}Mohamed~Cheriet} \\
	Department of Automated Manufacturing \\ Engineering\\
	\'Ecole de Technologie Sup\'erieure, \\ Universit\'e du Qu\'ebec\\
	Montr\'eal, QC, Canada \\
	\And
	\href{https://orcid.org/0000-0001-9484-7218}{\includegraphics[scale=0.06]{orcid.pdf}\hspace{1mm}Ghyslain~Gagnon} \\
	Department of Electrical Engineering\\
	\'Ecole de Technologie Sup\'erieure, \\ Universit\'e du Qu\'ebec\\
	Montr\'eal, QC, Canada \\
}

\renewcommand{\shorttitle}{Energy Disaggregation using Variational Autoencoders}

\hypersetup{
pdftitle={Energy Disaggregation using Variational Autoencoders},
pdfsubject={cs.LG},
pdfauthor={Antoine ~Langevin, Marc-André ~Carbonneau, Mohamed ~Cheriet, Ghyslain ~Gagnon},
pdfkeywords={Non-intrusive load monitoring (NILM), Energy Disaggregation, Variational Autoencoders (VAE), Generative Models},
}

\begin{document}
\maketitle

\begin{abstract}
	Non-intrusive load monitoring (NILM) is a technique that uses a single sensor to measure the total power consumption of a building. Using an energy disaggregation method, the consumption of individual appliances can be estimated from the aggregate measurement. Recent disaggregation algorithms have significantly improved the performance of NILM systems. However, the generalization capability of these methods to different houses as well as the disaggregation of multi-state appliances are still major challenges. In this paper we address these issues and propose an energy disaggregation approach based on the variational autoencoders framework. The probabilistic encoder makes this approach an efficient model for encoding information relevant to the reconstruction of the target appliance consumption. In particular, the proposed model accurately generates more complex load profiles, thus improving the power signal reconstruction of multi-state appliances. Moreover, its regularized latent space improves the generalization capabilities of the model across different houses. The proposed model is compared to state-of-the-art NILM approaches on the \mbox{UK-DALE} and REFIT datasets, and yields competitive results. The mean absolute error reduces by 18\% on average across all appliances compared to the state-of-the-art. The F1-Score increases by more than 11\%, showing improvements for the detection of the target appliance in the aggregate measurement.
\end{abstract}

\keywords{Non-intrusive load monitoring (NILM) \and Energy Disaggregation \and Variational Autoencoders (VAE) \and Generative Models
}

\section{Introduction}
\label{sec:introduction}
Since the building sector is responsible for 20\% of global energy consumption, energy conservation plans are introduced to encourage reductions in this sector \cite{0.1eia2016energy}. It is essential to consider smart energy management solutions to achieve the reduction goal \cite{7hosseini2017non}. However, these solutions require an appliance load monitoring system to obtain the status of individual appliances in households and to measure their energy consumption \cite{6welikala2019incorporating}.

Non-intrusive load monitoring (NILM) is a technique, proposed by Hart \cite{20hart1992nonintrusive}, that enables individual appliance load monitoring with a minimum of sensors. Usually, only one sensor measures the total power consumption of the building. An energy disaggregation algorithm is then used to separate individual appliance loads from the aggregate signal containing consumption of all appliances. While many solutions have been proposed, a few drawbacks still prevent the practical deployment of NILM: the scalability issue of NILM systems due to the complexity of the disaggregation task, which increases exponentially with the number of appliances and states, the challenge of near-real time capabilities for the disaggregation task, and model generalization are some examples of obstacles to \mbox{overcome \cite{19nalmpantis2019machine}.}

Traditional approaches for energy disaggregation include graph signal processing, Hidden Markov Models (HMM), and their variants \cite{17kolter2012approximate,18he2016non,16ji2019non}. However, they suffer from a scalability problem that hinders performance as the number of appliances increases. This situation is problematic in real use cases where more than 15 appliances can easily be present in a typical house.

Other approaches propose the training of one model per appliance type to alleviate the scalability issue. Generally, these approaches are based on deep neural networks (DNN), such as long-short term memory (LSTM) or denoising autoencoders (DAE) \cite{12kelly2015neural,mauch2015new,14bonfigli2018denoising,zhang2016sequencetopoint,jiang2019deep,kaselimi2019multi}. These solutions require sub-metering appliances during the supervised training phase.
These models disaggregate well when the training and testing data are extracted from the same house. However, they still need improvements to be as effective when deployed in a different house. Furthermore, some models, such as sequence-to-point (S2P) \cite{zhang2016sequencetopoint}, are non-causal approaches. Therefore, they cannot be used in a near-real time context, which is problematic for time-shifted strategies used by home energy management systems (HEMS) to schedule energy-consuming appliances. In this perspective, \cite{harell2019wavenilm} proposes an approach based on dilated convolutional layers for the task of energy disaggregation. This autoregressive network has the particularity of being causal. Therefore, it is more adapted to the disaggregation process without requiring a delay for the decision taker. Although this type of model gives promising results, it is more efficient when the number of appliances in the system remains low.

Another challenge in this field is that the performance of many approaches degrades significantly for multi-state appliances \cite{12kelly2015neural,kong2019practical,wu2019concatenate}. These multi-state appliances, such as washing machines and dishwashers, are among the appliances that account for a significant share of total consumption. Moreover, they are in the category of delayable appliances, making them strategic appliances that can be time-shifted by a HEMS within smart grids. Thus, NILM systems must be effective to disaggregate loads for this type of appliance.

Generative Adversarial Networks (GAN) \cite{goodfellow2014generative} have demonstrated impressive capabilities for image synthesis and other tasks in a variety of fields, such as video games. Particularly, \cite{bao2018enhancing, pan2020sequence} propose approaches based on the GAN framework for the energy disaggregation. The results show improved performance for the reconstruction of consumption signals, especially for multi-state appliances.

Variational autoencoders (VAE) \cite{VAE} are another important group of generative models that have received particular attention in recent years \cite{VAE_wavenet, liang2018variational}. As a generative model, the VAE is more stable during training. Its regularized latent space allows interpolations between two distributions learned during training to generate realistic appliance load profiles once decoded. Load profiles often vary from one activation to another for the same type of appliance. VAE is, therefore, well suited to the task of energy disaggregation, to generate the power signal of the target appliance using the aggregate power as input. Sirojan \textit{et al.} introduce the convolutional VAE for the energy disaggregation in \cite{sirojan2018deep}. Their method outperforms state-of-the-art approaches based on neural networks \cite{12kelly2015neural, zhang2016sequencetopoint}. This VAE framework demonstrates a promising avenue for generative approaches in the NILM domain.

In this work, we propose a method based on the VAE neural network framework for the energy disaggregation task. As in \cite{sirojan2018deep}, the proposed approach consists of two components: an encoder which maps information into a latent space, and a decoder that reconstructs the power signal of the target appliance using the latent representation. However, the proposed encoder uses a succession of instance-batch normalization networks \mbox{(IBN-Net)} \cite{xingangTwo} to enhance high-level feature extraction from the aggregate measurement. Furthermore, we implement skip connections between the encoder and decoder, allowing the decoder to benefit the feature maps from the encoder, as in the \mbox{U-Net} architecture \cite{ronneberger2015u}. These connections offer a global insight of the aggregate power consumption to the decoder and therefore allow a better reconstruction of the target signal by the deconvolution layers.

One of the key benefits of the proposed approach is the regularized latent space provided by the VAE framework which facilitates the encoding of relevant features of the aggregate signal. While batch and instance normalization in the model helps to stabilize and improve the learning process, skip connections enhance the signal reconstruction performance. The proposed approach is extensively analyzed against state-of-the-art approaches. To assess generalization capability, tests are carried out on \ctony{different houses from the \mbox{UK-DALE} \cite{kelly2015uk} and REFIT \cite{murray2017electrical} datasets.} The obtained results show that, on average, the proposed VAE-NILM outperforms state-of-the-art approaches and demonstrates higher accuracy for signal reconstruction of multi-state appliances.

The rest of this paper is organized as follows. \mbox{Section \ref{sec:related}} gives a brief overview of DNN techniques for NILM. The proposed solution based on VAE for the energy disaggregation task is presented in Section \ref{sec:solution}. Section \ref{sec:Setup} describes the experiment setup, followed by results and discussions in Section \ref{sec:Results}. Finally, the paper concludes and projects future research in Section \ref{sec:Conclusion}.

\section{Background}
\label{sec:related}
NILM is a technique used to monitor the individual consumption of household appliances using a single sensor that measures the total power of the whole house. The NILM problem can be formulated as follows: Let $x(t)$ be the aggregate measurements containing power consumption of all appliances at time $t$. Generally, we suppose that $x(t)$ represents the active power. Then the aggregate signal can be expressed by:

\begin{equation}
    x(t)=\sum_{i=1}^{M} y_i(t) + \epsilon(t),
\end{equation}
\vspace{0.2cm}

\noindent
where $y_i(t)$ represents the power consumption of the appliance $i$, $M$ is the number of appliances and $\epsilon(t)$ is the measurement's noise. Using an energy disaggregation algorithm, the objective is to recover individual appliance power consumption given only the aggregate measurement.

DNN approaches have been applied to energy disaggregation for many years \cite{12kelly2015neural, mauch2015new}. These state-based approaches are mainly used for low-frequency \mbox{(\textless 1 Hz)} monitoring which requires lower-cost hardware. Methods based on recurrent neural networks \cite{12kelly2015neural}, such as LSTM \cite{kim2017nonintrusive, mauch2015new} or Gated Recurrent Unit \cite{krystalakos2018sliding}, have been primarily proposed as they are well suited for 1D time series data. In \cite{kaselimi2019bayesian}, the authors extend the RNN and propose a Bayesian optimized bidirectional LSTM model for NILM.

To improve feature extraction, architectures based on convolutional neural network (CNN) have been proposed for energy disaggregation. Reference \cite{14bonfigli2018denoising} proposes a DAE approach for NILM. In this context, the input signal corresponds to the aggregate consumption of all appliances and the desired output is the power consumption of a single target appliance. Thus, the power consumption of appliances other than the target appliance is considered as noise in the input signal. The work of Zhang \textit{et al.} \cite{zhang2016sequencetopoint} presents two approaches based on CNN: sequence-to-sequence (S2S) and S2P. The latter is trained to predict the target appliance's power consumption only for the midpoint of the input window. This allows the model to focus on predicting a single value rather than the entire sequence considering that the window edges are harder to predict. The S2P approach improves energy disaggregation performance, but it significantly increases the computational complexity of the method compared to S2S. Xia \textit{et al.} propose the D-ResNet model \cite{xia2019non}, which uses dilated convolutional layers with residual connections. The dilated convolutional layers allow a larger receptive field without increasing the model's complexity, whereas the residual connections facilitate the gradient flow during training. However, most of these DNN models still suffer from a lack of generalization, and the test performance is often sensitive to the training dataset. Consequently, the models have more difficulty to disaggregate loads when deployed in further houses \cite{15DBLP:journals/corr/FaustineMKM17}.

\begin{figure*}
 \centering
 \includegraphics[width=\linewidth]{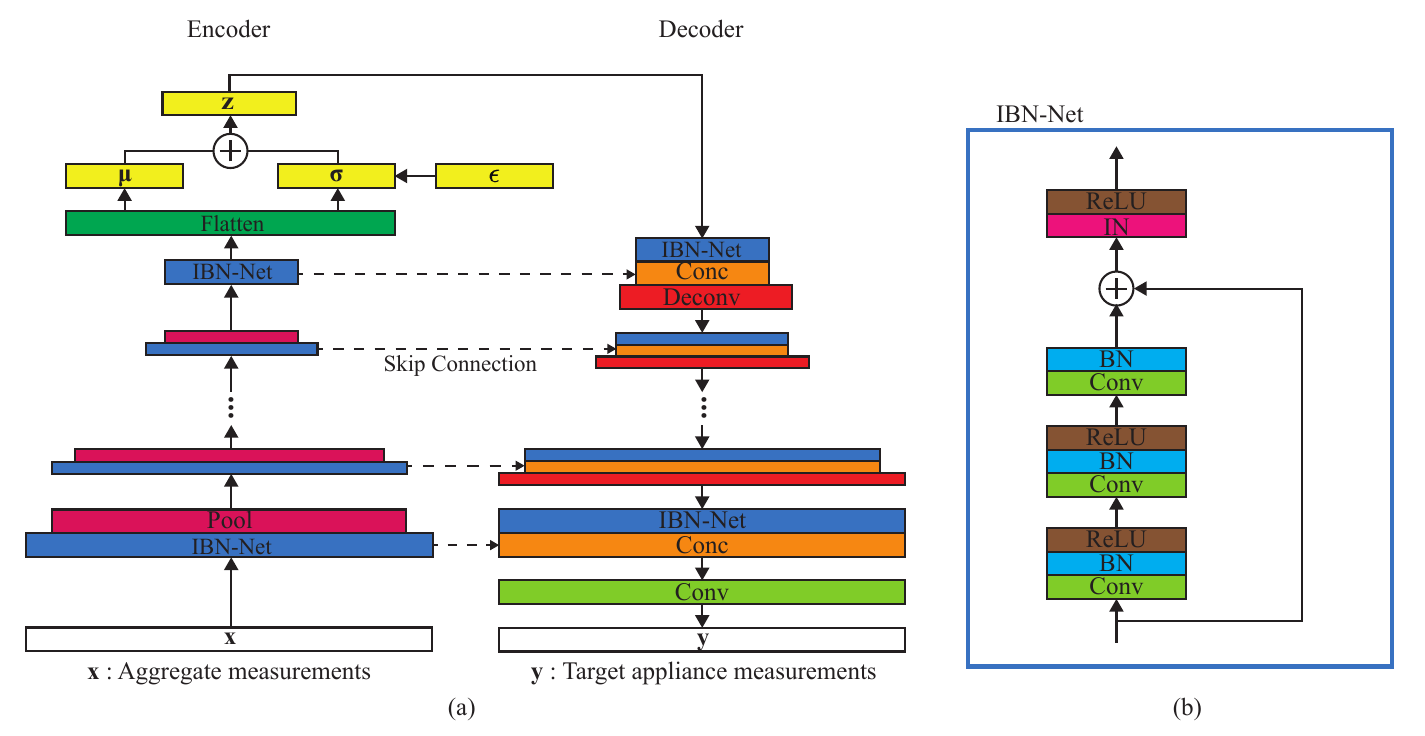}
 \caption{Proposed model with skip connections between layers of the encoder and the decoder (a), in addition a detailed view of the IBN-Net (b).}
 \vspace{-0.3cm}
 \label{fig:network_architecture}
\end{figure*}

Generative approaches for energy disaggregation, such as GAN \cite{bao2018enhancing, pan2020sequence} demonstrate improvements in signal reconstruction and generalization compared to state-of-the-art methods. In particularly, \cite{pan2020sequence} proposes a sequence-to-subsequence (S2SS) model to make a trade-off between traditional S2S and S2P methods, to balance the convergence difficulty and the computational complexity. The generator of the S2SS model is based on the U-Net architecture, which increases the quality of the generated power consumption signals. A generative model based on the VAE framework is proposed in \cite{sirojan2018deep}. The regularized latent space of the VAE framework encourages the encoder to map the relevant information contained in the aggregated signal. This allows the decoder to better reconstruct the power consumption signal of the target appliance, and thus offers better disaggregation performance than CNN-based models.

\section{Proposed Solution}
\label{sec:solution}
In this section, we first review the VAE framework and then describe the proposed model.

\subsection{VAE Framework}
The VAE \cite{VAE} proposes a probabilistic framework that maps inputs $\textbf{x} \in \mathbb{R}^{N}$ in a distribution over a continuous latent space $\textbf{z}$ rather than a single point. The true posterior density $p_{\theta}(\textbf{z}|\textbf{x})$ is intractable, which results in an indifferentiable marginal likelihood $p_{\theta}(\textbf{x})$. To address it, the probabilistic encoder $q_{\phi}(\textbf{z}|\textbf{x})$ is introduced to approximate the true conditional inference distribution $p_{\theta}(\textbf{z}|\textbf{x})$. The variational parameters $\phi$ are learned jointly with the generative model parameters $\theta$ following the variational principle, where $\text{log} \; p_{\theta}(\textbf{x})$ can be written as:

\vspace{-0.2cm}
{\setlength\arraycolsep{0.1em}
\begin{eqnarray}
\label{equ:VLB_Loss_function}
\text{log} \; p_{\theta}(\textbf{x})&=&\textbf{KL}\big(q_{\phi}(\textbf{z}|\textbf{x})||p_{\theta}(\textbf{z}|\textbf{x})\big) + \mathcal{L}(\theta, \phi; \textbf{x}) \nonumber \\
&\geq&\mathcal{L}(\theta, \phi; \textbf{x}) \nonumber \\
&=&\mathbb{E}_{q_{\phi}(\textbf{z}|\textbf{x})} \big[ \text{log}\; p_{\theta}(\textbf{x}|\textbf{z}) \big] -\textbf{KL}\big(q_{\phi}(\textbf{z}|\textbf{x})||p_{\theta}(\textbf{z})\big), \nonumber \\
\end{eqnarray}}
\vspace{-0.5cm}

\noindent
where $\mathcal{L}(\theta, \phi; \textbf{x})$ is the variational lower bound to optimize. The expectation term of (\ref{equ:VLB_Loss_function}) encourages the reconstruction accuracy of the probabilistic decoder $p_{\theta}(\textbf{x}|\textbf{z})$, while the Kullback-Leibler ($\textbf{KL}$) divergence term acts as a regularizer to constrain the approximate posterior to be close to the prior $p_{\theta}(\textbf{z})$. Generally, $p_{\theta}(\textbf{z})$ is assumed to be a centered isotropic multivariate Gaussian with identity covariance $\mathcal{N}(\textbf{z}; \textbf{0},\textbf{I})$ and the variational approximate posterior is $\mathcal{N}(\textbf{z}; \bm{\mu}(\textbf{x}), \bm{\sigma}^2(\textbf{x})\textbf{I})$. In this case, both $p_{\theta}(\textbf{z})$ and $q_{\phi}(\textbf{z}|\textbf{x})$ are Gaussian, so that $\textbf{KL}\big(q_{\phi}(\textbf{z}|\textbf{x})||p_{\theta}(\textbf{z})\big)$ can be calculated in closed form. In practice, both the encoder $q_{\phi}(\textbf{z}|\textbf{x})$ and the decoder $p_{\theta}(\textbf{x}|\textbf{z})$ are neural networks parameterized by $\phi$ and $\theta$ respectively. This way, $\bm{\mu}(\textbf{x})$ and $\bm{\sigma}^2(\textbf{x})$ correspond to the encoder outputs and they are learned from observed datasets through the objective function given in (\ref{equ:VLB_Loss_function}).

Since the decoder samples from $\textbf{z} \sim q_{\phi}(\textbf{z}|\textbf{x})$, the gradient cannot be backpropagated through the stochastic units within the network. To address it, we use a reparameterization trick \cite{VAE}. We start by sampling an auxiliary variable $\epsilon \sim \mathcal{N}(0, \textbf{I})$ and then compute $\textbf{z} = \bm{\mu} + \bm{\sigma} \odot \epsilon$. Here, $\odot$ denotes an element-wise product.

\subsection{Proposed Model}
In this work, we propose an energy disaggregation model based on the VAE \cite{VAE} framework. The whole network consists of two components, as shown in Fig. \ref{fig:network_architecture} (a): The encoder distills relevant target appliance information from the aggregate signal $\textbf{x}$ into the latent space $\textbf{z}$, and the decoder reconstructs only the power signal of the target appliance from $\textbf{z}$.

The inputs of the model are sequences of aggregate power extracted containing $T$ time steps. The sequences are obtained using a sliding window. Each input sequence $\textbf{x} = [x_1, x_2, ..., x_T]$ is processed separately by the model to generate an output sequence $\textbf{y}= [y_1, y_2, ..., y_T]$ corresponding to the power of the target appliance. 

The proposed network architecture is composed of IBN-Nets as shown in Fig. \ref{fig:network_architecture} (b). The \mbox{IBN-Net} subnetwork architecture combines instance and batch normalization \cite{xingangTwo}. Batch normalization in convolutional layers increases the discriminating capacity of the learned features and thus allows the encoder to have more relevant features to map on the latent space. Moreover, instance normalization in shallow layers of the network improves generalization performance, which remains one of the weaknesses of many NILM approaches. The \mbox{IBN-Net} consists of three successive convolution layers combined with batch normalization and a rectified linear unit activation function (ReLU). A residual connection joins the \mbox{IBN-Net} input to the instance normalization layer to facilitate gradient flow through the entire model during training and also prevents the vanishing gradient problem.

The encoder architecture comprises seven \mbox{IBN-Nets} each followed by a max-pooling layer to decrease the temporal resolution. This encourages the learning of high-level features describing the target appliance. Two fully connected layers translate the output of the \mbox{IBN-Net} stack into the distribution parameters $\bm{\mu}$ and $\bm{\sigma}$ of the latent space $\textbf{z}$.

The architecture of the decoder is similar to that of the encoder. It consists of seven \mbox{IBN-Nets} followed by deconvolution layers to progressively increase the temporal resolution and reconstruct the signal of the target appliance. We concatenate through skip connections the outputs of the corresponding \mbox{IBN-Net} from the encoder to the decoder. In this way, the decoder can also benefit from the feature maps extracted by the encoder, useful for recovering the power consumption details from the aggregate measurement. The reconstruction of the target appliance consumption is thus more accurate than if it uses only the latent representation.

\section{Experiment}
\label{sec:Setup}
We experiment to assess the proposed \mbox{VAE-NILM} performance against state-of-the-art methods. We design the experiment to measure the detection and reconstruction capabilities of appliance power consumption. We divide the experimental protocol into three use-case scenarios that involve evaluating the model on three different houses \ctony{on the \mbox{UK-DALE} \cite{kelly2015uk} dataset and one use-case scenario on the REFIT \cite{murray2017electrical} dataset.}

\begin{table*}[t]
	\caption{Houses distribution for the use-case scenarios of the \mbox{UK-DALE} and REFIT datasets.}
	\resizebox{\textwidth}{!}{%
		\begin{tabular}{cclclccclclclclc}
			\cmidrule[1.5pt]{1-6}
			\cmidrule[1.5pt]{8-16}
			\textbf{Set}   & \multicolumn{5}{c}{\textbf{UK-DALE}} & \textbf{}         & \multicolumn{9}{c}{\textbf{REFIT}}                                 \\
			\cmidrule[1.5pt]{1-6}
			\cmidrule[1.5pt]{8-16}
			\multicolumn{1}{l}{} & \multicolumn{1}{l}{} &  & \multicolumn{1}{l}{} &  & \multicolumn{1}{l}{} & & \multicolumn{1}{l}{} &  & \multicolumn{1}{l}{} &  & \multicolumn{1}{l}{} &  &  &  &  \\[-2ex]
			\textbf{}            & \textbf{Scenario 1}  &  & \textbf{Scenario 2}  &  & \textbf{Scenario 3} & & \textbf{Fridge}      &  & \textbf{Kettle}                     &  & \textbf{Microwave}   &  & \multicolumn{1}{c}{\textbf{Washing M.}} &  & \multicolumn{1}{c}{\textbf{Dishwasher}} \\ \cline{2-2} \cline{4-4} \cline{6-6} \cline{8-8} \cline{10-10} \cline{12-12} \cline{14-14} \cline{16-16}
			\multicolumn{1}{l}{} & \multicolumn{1}{l}{} &  & \multicolumn{1}{l}{} &  & \multicolumn{1}{l}{} & \multicolumn{1}{l}{} & \multicolumn{1}{l}{} &  & \multicolumn{1}{l}{}                &  & \multicolumn{1}{l}{} &  &                                             &  &                                         \\[-1.5ex]
			\textbf{Train}       & 1, 5                 &  & 2, 5                 &  & 1, 2                 &	\textbf{}       & 2, 5, 9, 12          &  & \begin{tabular}[c]{@{}c@{}}3, 4, 5, 6, 7, 8,\\9, 12, 13, 19, 20\end{tabular} &  & 10, 12, 17, 19       &  & \begin{tabular}[c]{@{}c@{}}2, 5, 7, 9, 15,\\16, 17, 18\end{tabular} &  & \begin{tabular}[c]{@{}c@{}}5, 7, 9, 13,\\16, 18\end{tabular} \\
			\multicolumn{1}{l}{} & \multicolumn{1}{l}{} &  & \multicolumn{1}{l}{} &  & \multicolumn{1}{l}{} & \multicolumn{1}{l}{} & \multicolumn{1}{l}{} &  & \multicolumn{1}{l}{}                &  & \multicolumn{1}{l}{} &  &                                             &  &                                         \\[-2ex]
			\textbf{Test}        & 2                    &  & 1                    &  & 5                    & \textbf{}        & 15                   &  & 2                                   &  & 4                    &  & \multicolumn{1}{c}{8}                       &  & \multicolumn{1}{c}{20}                  \\
			\cmidrule[1.5pt]{1-6}
			\cmidrule[1.5pt]{8-16}
			\multicolumn{1}{l}{} & \multicolumn{1}{l}{} &  & \multicolumn{1}{l}{} &  & \multicolumn{1}{l}{} & \multicolumn{1}{l}{} & \multicolumn{1}{l}{} &  & \multicolumn{1}{l}{}                &  & \multicolumn{1}{l}{} &  &                                             &  &                                         \\[-1.5ex]
			\textbf{}   & \multicolumn{5}{c}{{\large(a)}} & \textbf{}         & \multicolumn{9}{c}{{\large(b)}}                                 \\
	\end{tabular}}
	\vspace{-0.5cm}
	\label{tab:appliance_carac_a_b}
\end{table*}

\subsection{\ctony{Datasets}}
The \mbox{UK-DALE} dataset is used in our \ctony{first three} scenarios. This reference dataset is a set of five houses containing respectively, 54, 20, 5, 6, and 26 sub-metered appliances. Each house includes the aggregate measurement of electrical power as well as the individual electrical power of all appliances sampled at $1/6$ Hz. In this paper, we focus the disaggregation on three On/Off appliances: refrigerator, kettle and microwave, and on two multi-states appliances: washing machine and dishwasher which are all available in houses 1, 2, and 5. We ignore houses 3 and 4 because they contain only a few appliances.

\ctony{The REFIT dataset is a set of 20 houses which contain on average nine sub-metered appliances each. The aggregate and sub-metered measurements of the active power are both sampled at $1/8$ Hz. The same five appliances chose for experiments on the \mbox{UK-DALE} dataset are selected for the REFIT scenario. During data preparation, several missing data were detected in the aggregate and sub-metered readings. Therefore, data preprocessing was performed to remove missing data longer than five minutes. However, missing data of a few seconds to five minutes were kept because they could be considered as noise. They also reproduce small data losses that could be related to a loss of communication during deployment.} 

\subsection{Reference Methods}
\label{subsec:Ref_methods}
We compare our model against four state-of-the-art methods. We select the DAE approach proposed by Kelly \textit{et al.} \cite{12kelly2015neural} as a first comparison method. Second, we compare the proposed model with the work of Zhang \textit{et al.} \cite{zhang2016sequencetopoint}, which proposes two approaches, S2S and S2P, based on CNN. Finally, we choose the generative approach, S2SS \cite{pan2020sequence}, based on the GAN framework as a comparison algorithm. We carry out experiments using the original authors' implementations available for each method: DAE\footnote{https://github.com/JackKelly/neuralnilm}, S2S\footnote{https://github.com/MingjunZhong/NeuralNetNilm} and S2P\footnote{https://github.com/MingjunZhong/seq2point-nilm}, and S2SS\footnote{https://github.com/DLZRMR/seq2subseq}.

\subsection{Experimental Protocol}
In this work, we focus on scenarios where test houses are held out during the training process, as is done in \cite{12kelly2015neural, zhang2016sequencetopoint, pan2020sequence}. For each scenario \ctony{using the \mbox{UK-DALE} data}, we hold out a house for the test and train the model on the other two houses, as detailed in \mbox{Table \ref{tab:appliance_carac_a_b}} (a). The data available for \ctony{the \mbox{UK-DALE}} house 1 correspond to a collection period of more than five years, which is much longer than for the other houses. Thus, to balance the training set, we randomly select 15\% of the total number of sequences in house 1 and we complete the training set with all sequences of a second house. \ctony{For the fourth scenario based on the REFIT dataset, we hold out one house per appliance and train the models on the other selected houses, as detailed in Table \ref{tab:appliance_carac_a_b} (b).} In this way, each experiment represents a real use case where the model would be deployed in houses without the need for sub-metering during a training period.

\ctony{For each scenario,} we train one model for each type of appliance on 80\% of the training set, and we use the remaining 20\% for validation. For each house, the aggregate and sub-metered measurements are separated into sequences using a sliding window. We employ the same 
technique for the test house and recombine overlapped portions using the median filter, as done in \cite{14bonfigli2018denoising}. We reproduce the same protocol for each reference method. Finally, we repeat the process ten times and average scores over all repetitions. We use Welch's t-test to assess the statistical significance of the results.

\subsection{Implementation Details}
The encoder and decoder of the proposed model contain seven groups of layers, i.e., \mbox{IBN-Net} and max pooling for the encoder, and \mbox{IBN-Net}, concatenation, and deconvolution for the decoder. The number of filters for the convolution layers in all \mbox{IBN-Nets} is 64, 64, and 256 respectively. The max-pooling divides by two the temporal resolution every step in the encoder, whereas the deconvolution operation increases by two the temporal resolution in the decoder. All the hyperparameters of the proposed model and training process were tuned during the preliminary experiments using a grid search technique.

We train the proposed model through supervised learning using the objective function (\ref{equ:VLB_Loss_function}). We use the optimizer RMSProp (Root Mean Square Propagation) with a decreasing learning rate initialized at 0.001, which decreases by half at each epoch. All experiments are run on a maximum of 100 epochs using an early-stopping criterion with a patience of 20 epochs to prevent overfitting.

These reference methods DAE, S2S, and S2P adapt the input size $T$ to the appliance type on the basis of the average activation time. In practice, they extract all appliance activations from the sub-metered measurements of the training houses, and they calculate the average activation time for each type of appliance. However, a larger window is preferable, even for appliances with short activation duration, such as the kettle and microwave. This gives the model a better overview of other active loads, and it better discerns activations related to the target appliance. The proposed model and S2SS use the same size $T$ set to 1024 for all appliance types. Thus, $T$ is large enough to contain the full activation of a multi-state appliance. All hyperparameters are fixed regardless of the appliance type, except for the stride $S$ of the sliding window and the batch size. For both multi-state appliances, the stride $S$ and the batch size are set to 256 and 32 respectively, and they are set to 64 and 150 for the On/Off appliances. The code used to create results in this paper is available in our "\mbox{VAE-NILM}" repository.\footnote{https://github.com/ETSSmartRes/\mbox{VAE-NILM}} 

The reference methods are trained according to the implementation guidelines provided in the repositories listed in Section \ref{subsec:Ref_methods}. However, we adjusted some hyperparameters, such as batch size and window size, to improve their performances. Furthermore, we added dropouts in models S2S and S2P to reduce overfitting.

\begin{table*}[t]
	\caption{Results for models trained on houses 1 and 5 from the UK-DALE dataset and tested on the house 2.}
	\resizebox{\linewidth}{!}{
		\begin{tabular}{llllllll} 
			\toprule[2pt]
			&           & Fridge                     & Kettle                    & Microwave                  & Washing M.                & Dishwasher                &                         \\ 
			Metric & Method    & \textit{On/Off}            & \textit{On/Off}           & \textit{On/Off}            & \textit{Multi-State}      & \textit{Multi-State}      & Average                 \\ 
			\bottomrule[2pt]                                                                                       
			&           &                           &                           &                           &                           &                           &                         \\[-2ex]
			MAE                & DAE       & $21.8\pm0.8$              & $11.8\pm1.1$              & $10.2\pm1.2$              & $22.3\pm3.1$              & $42.9\pm13.7$             & $21.8\pm4.0$            \\
			(W)                & S2S       & $21.9\pm1.1$              & $13.6\pm0.8$              & $10.9\pm1.3$              & $11.3\pm0.8$              & $25.4\pm6.2$              & $16.6\pm2.0$            \\
			& S2P       & $16.9\pm0.6$              & $9.0\pm1.2$               & $15.0\pm4.8$              & $13.3\pm2.8$              & $19.2\pm7.4$              & $14.7\pm3.3$            \\
			& S2SS      & $19.7\pm0.4$              & $12.8\pm0.5$              & $19.1\pm0.6$              & $9.1\pm1.4$               & $18.1\pm2.3$              & $15.8\pm1.0$            \\
			& VAE-NILM  & \boldmath{$15.1\pm0.3$}   & \boldmath{$6.1\pm0.5$}    & \boldmath{$5.1\pm0.2$}    & \boldmath{$6.2\pm0.2$}    & \boldmath{$11.6\pm3.0$}   & \boldmath{$8.8\pm0.8$}  \\
			&           &                           &                           &                           &                           &                           &                         \\[-2ex]
			$\text{MAE}_{ON}$  & DAE       & $28.4\pm1.8$  & $1045.1\pm47$            & $1128.7\pm29$           & $579.4\pm18$            & $1234.1\pm56$           & $803.2\pm21$          \\
			(W)                & S2S       & $28.3\pm2.3$              & $1271.5\pm88$           & $1247.2\pm49$            & $447.8\pm7$             & $676.3\pm75$            & $734.2\pm34$          \\
			& S2P       & $19.3\pm1.1$              & $745.0\pm117$           & $1148.4\pm50$           & $458.5\pm138$           & $605.7\pm268$           & $595.4\pm115$         \\
			& S2SS      & $21.0\pm0.6$              & $1150.0\pm52$           & $1186.8\pm48$           & $254.1\pm38$            & $466.8\pm61$            & $615.7\pm40$          \\
			& VAE-NILM  & \boldmath{$14.8\pm0.8$}   & \boldmath{$478.1\pm46$} & \boldmath{$788.2\pm26$} & \boldmath{$215.1\pm17$} & \boldmath{$351.0\pm107$}& \boldmath{$369.4\pm39$}\\
			&           &                           &                           &                           &                           &                           &                         \\[-2ex]
			EpD    & DAE      & $242\pm22$                & $236\pm21$                 & \boldmath{$77\pm6$}       & $291\pm57$                & $738\pm177$               & $317\pm52$              \\
			(Wh)   & S2S      & $250\pm23$                & $287\pm28$                & $82\pm10$                  & $141\pm16$                & $384\pm77$                & $229\pm29$              \\
			& S2P      & \boldmath{$174\pm31$}     & $179\pm28$                & $151\pm63$                & $153\pm38$                & $402\pm198$               & \boldmath{$212\pm72$}   \\
			& S2SS     & \boldmath{$161\pm13$}     & $283\pm12$                & $175\pm19$                & \boldmath{$100\pm21$}     & $279\pm35$                & $200\pm20$              \\
			& VAE-NILM & \boldmath{$168\pm18$}     & \boldmath{$120\pm12$}     & \boldmath{$78\pm4$}       & \boldmath{$111\pm6$}      & \boldmath{$200\pm80$}     & \boldmath{$135\pm24$}   \\
			&          &                           &                           &                           &                           &                           &                         \\[-2ex]
			PR     & DAE      & $90.7\pm1.1$              & $78.5\pm3.2$              & $57.4\pm9.7$              & $13.5\pm4.3$              & $78.0\pm14.6$             & $63.6\pm6.6$            \\
			(\%)   & S2S      & \boldmath{$93.6\pm0.8$}   & $77.5\pm1.2$              & $35.2\pm17.4$             & $37.0\pm8.1$              & $69.5\pm17.2$             & $62.6\pm8.9$            \\
			& S2P      & $91.9\pm1.2$              & $89.5\pm1.7$              & $37.7\pm20.8$             & $32.1\pm10.3$             & \boldmath{$91.3\pm4.9$}   & $68.5\pm7.8$            \\
			& S2SS     & $85.4\pm0.8$              & $93.9\pm0.7$              & $63.6\pm15.6$             & $70.1\pm6.9$              & $75.6\pm5.5$              & $77.7\pm5.9$            \\
			& VAE-NILM & $91.2\pm0.9$              & \boldmath{$96.7\pm0.6$}   & \boldmath{$83.1\pm3.1$}   & \boldmath{$85.8\pm1.5$}   & \boldmath{$94.7\pm2.8$}   & \boldmath{$90.3\pm1.8$} \\
			&          &                           &                           &                           &                           &                           &                         \\[-2ex]
			RE     & DAE      & $70.2\pm3.2$              & $99.7\pm0.1$              & $20.9\pm5.1$              & $80.1\pm9.6$              & \boldmath{$70.7\pm11.2$}  & $68.3\pm5.9$            \\
			(\%)   & S2S      & $71.7\pm4.1$              & \boldmath{$99.9\pm0.1$}   & \boldmath{$42.8\pm23.1$}  & \boldmath{$99.2\pm0.4$}   & \boldmath{$74.4\pm24.8$}  & \boldmath{$77.6\pm10.5$}\\
			& S2P      & $83.3\pm1.7$              & $99.1\pm0.5$              & $20.1\pm9.0$              & $92.6\pm7.1$              & \boldmath{$68.5\pm13.5$}  & $72.7\pm6.4$            \\
			& S2SS     & $82.2\pm0.7$              & $94.6\pm1.1$              & $10.8\pm6.3$              & $93.3\pm1.9$              & \boldmath{$76.8\pm4.9$}   & $71.5\pm3.0$            \\
			& VAE-NILM & \boldmath{$87.6\pm1.1$}   & $97.9\pm0.4$              & \boldmath{$46.4\pm2.7$}   & $94.9\pm0.6$              & \boldmath{$75.8\pm8.7$}   & \boldmath{$80.5\pm2.7$} \\
			&          &                           &                           &                           &                           &                           &                         \\[-2ex]
			F1     & DAE      & $79.1\pm1.7$              & $87.8\pm2.1$              & $30.2\pm5.8$              & $22.6\pm7.0$              & $72.0\pm7.2$              & $58.4\pm4.8$            \\
			(\%)   & S2S      & $81.2\pm2.8$              & $87.2\pm0.8$              & $36.5\pm17.3$             & $53.3\pm8.0$              & \boldmath{$70.9\pm23.5$}  & $65.8\pm10.5$           \\
			& S2P      & $87.4\pm1.0$              & $94.0\pm1.0$              & $20.6\pm7.7$              & $46.4\pm10.6$             & \boldmath{$77.1\pm11.2$}  & $65.1\pm6.3$            \\
			& S2SS     & $83.7\pm0.5$              & $94.2\pm0.6$              & $17.0\pm9.2$              & $79.8\pm4.2$              & $76.0\pm3.6$              & $70.1\pm3.6$            \\
			& VAE-NILM & \boldmath{$89.3\pm0.5$}   & \boldmath{$97.3\pm0.2$}   & \boldmath{$59.5\pm2.8$}   & \boldmath{$90.1\pm0.8$}   & \boldmath{$83.8\pm4.8$}   & \boldmath{$84.0\pm1.8$} \\
			&          &                           &                           &                           &                           &                           &                         \\[-2ex]
			\bottomrule[2pt]
		\end{tabular}
	}
	\label{tab:Result_2}
\end{table*}

\subsection{Performance Metrics}
To compare our model with the state-of-the-art, we evaluate the disaggregation performance using several metrics. First, we compute the mean absolute error (MAE) between the predicted and ground truth power:

\begin{equation}
\label{equ:MAE}
\textit{MAE} = \frac{1}{T} \sum_{t=1}^{T} |\hat{y}_t-y_t|,
\end{equation}
\vspace{0.2cm}

\noindent
where $T$ is the number of time points, $\hat{y}_t$ and $y_t$ are respectively the predicted power and the ground truth power at time $t$.

\begin{figure*}
	\centering
	\includegraphics[width=1\linewidth]{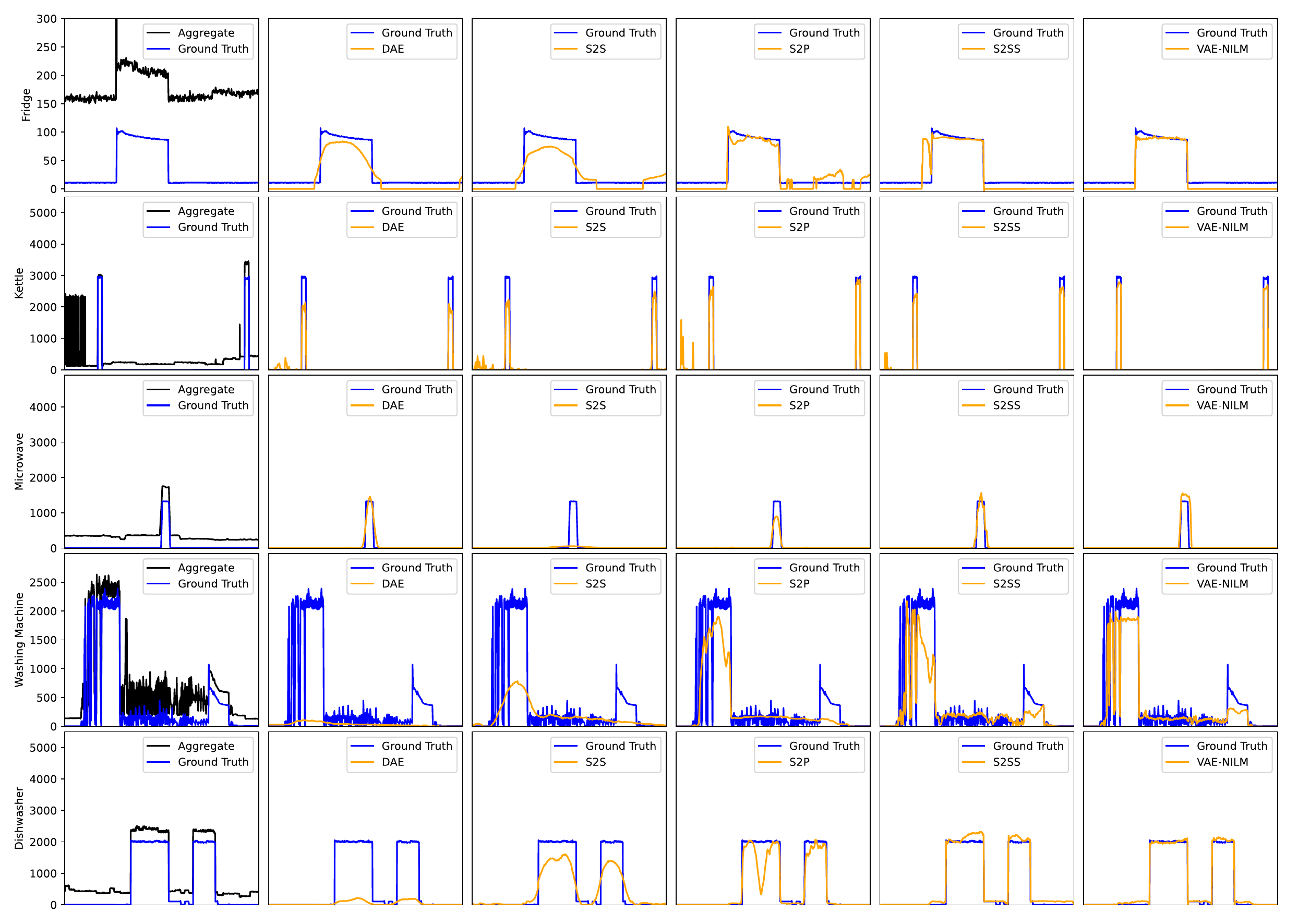}
	\caption{Examples of disaggregation results on UK-DALE house 2, between the proposed VAE-NILM model and all reference methods for the five appliances.}
	\vspace{-0.5cm}
	\label{fig:Activations}
\end{figure*}

Particularly useful for household users, energy per day (EpD) \cite{d2019transfer} is a measure that calculates the absolute error of the predicted energy in a one-day period. The average EpD over the entire dataset is defined as follows:

\begin{equation}
\label{equ:EpD}
\textit{EpD} = \frac{1}{D} \sum^{D}_{d=1} |\hat{e}-e|,
\end{equation}
\vspace{0.2cm}

\noindent
where $D$ is the total number of days and $e=\sum_{t} y_t$ is the energy consumption in a one-day period. 

In addition to metrics focused on energy consumption, we use state-based metrics to measure the ability of the model to predict the appliance state (ON/OFF). At each time step, an appliance is considered in the ON state if its power exceeds a predefined threshold \cite{12kelly2015neural}. The threshold $\delta$ varies depending on the appliance type. It corresponds to 50, 2000, 200, 20, and 10 W for the fridge, kettle, microwave, washing machine, and dishwasher respectively. Once states are assigned, we compute the F1-Score using the precision and recall measures.

Since appliances are mostly inactive, we define an additional metric, $\text{MAE}_{ON}$, that use the threshold $\delta$ to calculate the MAE separately only when the appliance is ON:

\begin{equation}
\label{equ:MAE_ON}
\textit{MAE}_{ON} = \frac{1}{N_{ON}} \sum_{}^{} |\hat{y}_t-y_t| \; \forall \; t \; \text{when} \; y_t \geq \delta,
\end{equation}
\vspace{0.2cm}

\noindent
where $t \in [1, ..., T]$ and $N_{ON}$ is the number of time points when the appliance is ON. $\text{MAE}_{ON}$ provides a more insightful evaluation of the algorithm accuracy, since the error is not averaged out during the time the appliance is actually off.

\section{Results and Discussions}
\label{sec:Results}
To facilitate comparison with state-of-the-art works, Table \ref{tab:Result_2} presents the test results for the energy disaggregation on house 2 with models trained on houses 1 and 5. The first two columns represent the metrics and methods respectively. The next five columns correspond to the specific results for each appliance. Finally, the last column is the average result for all appliances. Each reported result corresponds to the mean performance and the standard deviation for the ten repetitions. Bold values represent the optimal result of a metric under different methods. The results show that \mbox{VAE-NILM} yields the best results on average for all the metrics. When compared to the reference methods, we note that \mbox{VAE-NILM} reduces the MAE by 40\% on average over the entire test sequence with an average reduction of 226 W when appliances are ON. Moreover, the estimated EpD decreases by 32\% on average, while the F1-Score increases by 14\% in comparison with the other methods. We also observe that the proposed model improves the disaggregation performance evenly for multi-state appliances as well as appliances with short activation times. 

To reflect the overall performance on the \mbox{UK-DALE} dataset, Table \ref{tab:Result_all} presents the combined results of all the scenarios performed on houses 1, 2 and 5. Overall, we observe a trend similar to that of \mbox{house 2}. The MAE improves by about 18\% with a reduction of 130 W for the $\text{MAE}_{ON}$, whereas the EpD decreases by 13\%, and the \mbox{F1-Score} increases by 11\% compared to the state-of-the-art.

\ctony{Table \ref{tab:Result_REFIT} reports the results for the use-case scenario performed on the test houses of the REFIT dataset. In general, we observe similar findings to those obtained on the \mbox{UK-DALE} dataset. On average, the MAE improves by about 33\% with an average reduction of 77 W compared to the S2P method for $\text{MAE}_{ON}$. The EpD is 63 Wh lower, while the prediction, recall and \mbox{F1-Score} all increased by 12\% on average for the proposed model.}

\subsection{Disaggregated signal qualitative analysis}
Fig. \ref{fig:Activations} shows examples of disaggregated signals from house 2 for each type of appliance. The first column shows the aggregate signal and the ground truth signal of the target appliance. The following columns illustrate the ground truth signal and the predicted signal for each reference model, and the last column depicts the proposed model. Better $\text{MAE}_{ON}$ performance could be explained by a more accurate signal reconstruction of the \mbox{VAE-NILM} than reference methods. In particular, the proposed model improves reconstruction performance for multi-state appliances that have a longer activation time, as shown in \mbox{Fig. \ref{fig:Activations}.}

\begin{table}[t]
	\centering
	\caption{Results of the overall experiment combining the three test scenarios on houses 1, 2, 5 of the \mbox{UK-DALE} dataset.}
	\vspace{0.1cm}
	\resizebox{0.7\columnwidth}{!}{
		\begin{tabular}{llllllll} 
			\toprule[2pt]
			Metric 		& Method   		& Fridge           & Kettle            & Micro             & Wash              & Dish              & Average                 \\ 
			\bottomrule[2pt]                               
			&		        &		           &		           &		           &		           &		           &		    \\
			MAE    	    &	 DAE      	&	29.5	       &	32.0	       &	25.9	       &	11.0	       &	\textbf{24.8}  &	24.6	\\
			(W)    	    &	 S2S      	&	32.7	       &	26.2	       &	19.6	       &	12.2	       &	27.3	       &	23.6	\\
			&	 S2P      	&	28.2	       &	30.7	       &	18.5	       &	9.4	           &	29.4	       &	23.3	\\
			&	 S2SS     	&	\textbf{21.8}  &	24.7	       &	15.4	       &	12.2	       &	29.0	       &	20.6	\\
			&	 VAE-NILM 	&	\textbf{21.6}  &	\textbf{22.1}  &	\textbf{10.8}  &	\textbf{6.7}   &	\textbf{23.4}  &	\textbf{16.9}	\\
			&	          	&		           &		           &		           &		           &		           &		    \\[-2ex]
			$\text{MAE}_{ON}$ 	&	 DAE      	&	33.9	           &	543.9	           &	986.6	           &	908.8	           &	1211.2	       &	736.9	    \\
			(W)    	    &	 S2S      	&	36.5	           &	450.6	           &	514.1	           &	1288.2	       &	1169.8	       &	691.9	    \\
			&	 S2P      	&	30.4	           &	468.2	           &	\textbf{487.7}   &	1050.5	       &	1073.6	       &	622.1	    \\
			&	 S2SS     	&	28.1	           &	\textbf{390.8}   &	563.2	           &	902.1	           &	1254.6	       &	627.7	    \\
			&	 VAE-NILM 	&	\textbf{23.4}	   &	\textbf{382.4}   &	\textbf{441.5}   &	\textbf{733.4}   &	\textbf{881.5}   &	\textbf{492.5}	    \\
			&	          	&		           &		           &		           &	    	       &		           &		    \\[-2ex]
			EpD    	    &	 DAE      	&	189	           &	485	           &	432	           &	148	           &	449	           &	341	    \\
			(Wh)   	    &	 S2S      	&	219	           &	\textbf{368}   &	297	           &	200	           &	\textbf{386}   &	294	    \\
			&	 S2P      	&	196	           &	\textbf{397}   &	337	           &	160	           &	\textbf{423}   &	302	    \\
			&	 S2SS     	&	\textbf{128}   &	\textbf{360}   &	256	           &	208	           &	471	           &	\textbf{284}	    \\
			&	 VAE-NILM 	&	170	           &	\textbf{399}   &	\textbf{187}   &	\textbf{96}    &	\textbf{400}   &	\textbf{250}	    \\
			&	          	&		           &		           &		           &		           &		           &		    \\[-2ex]
			F1     	    &	 DAE      	&	78.0	       &	37.6	       &	53.8	       &	62.5	       &	19.3	       &	50.3	\\
			(\%)   	    &	 S2S      	&	75.5	       &	53.1	       &	53.7	       &	60.9	       &	22.3	       &	53.1	\\
			&	 S2P      	&	77.6	       &	50.9	       &	55.4	       &	72.1	       &	17.1	       &	54.6	\\
			&	 S2SS     	&	\textbf{80.0}  &	63.8	       &	54.8	       &	76.3	       &	8.9	           &	56.8	\\
			&	 VAE-NILM 	&	\textbf{80.6}  &	\textbf{73.5}  &	\textbf{64.6}  &	\textbf{87.1}  &	\textbf{32.1}  &	\textbf{67.6}	\\
			&	        	&	     	       &	     	       &	    	       &	     	       &	    	       &	     	\\[-2ex]
			\bottomrule[2pt]
		\end{tabular}
	}
	\vspace{0.3cm}
	\label{tab:Result_all}
\end{table}

The U-shaped architecture of the proposed model allows the feature maps extracted by the encoder to be combined with the deconvolution layers in the decoder through skip connections. The decoder then learns to assemble a more precise output based on this information. Thus, we note that the \mbox{VAE-NILM} model increases sharpness and accuracy of the reconstruction signals. While skip connections are also used in S2SS, we find that both models reconstruct multi-state appliances better than state-of-the-art.

Furthermore, we notice in Fig. \ref{fig:Activations} that the activation profile of the washing machine in house 2 starts with cyclic states, which is not the case for the washing machines in training houses 1 and 5. Despite this, we observe that the \mbox{VAE-NILM} and S2SS reconstruct the cyclic states, whereas other reference methods experience difficulties. We therefore conclude that skip connections provide valuable information to the decoder, making it more efficient in reconstructing the power consumption of the target appliance.

\begin{table}[t]
	\centering
	\caption{\ctony{Disaggregation results on the test houses of the REFIT dataset.}}
	\vspace{0.1cm}
	\resizebox{0.7\columnwidth}{!}{
		\begin{tabular}{llllllll} 
			\toprule[2pt]
			Metric 		& Method   		& Fridge           & Kettle            & Micro             & Wash              & Dish              & Average                 \\ 
			\bottomrule[2pt]                               
			&		        &		           &		           &		           &		           &		           &		    \\
			MAE                & DAE       & 19.0              & 14.4             & 8.6              & 34.3              & 12.1             & 17.7           \\
			(W)                & S2S       & 23.4              & 20.0             & 9.4              & 30.7              & 11.4             & 19.0           \\
			& S2P       & 20.3              & 12.3             & 7.0              & 21.9              & 11.7             & 14.7           \\
			& S2SS      & 20.0              & 8.7              & 9.9              & 15.7              & 6.2              & 12.1           \\
			& VAE-NILM  & \textbf{11.6}     & \textbf{6.4}     & \textbf{6.0}     & \textbf{12.0}     & \textbf{4.2}     & \textbf{8.0}   \\
			&           &                   &                  &                  &                   &                  &                \\[-2ex]
			$\text{MAE}_{ON}$  & DAE       & \textbf{22.8}     & 413.4           & 610.7            & 425.1             & 263.1            & 347.0          \\
			(W)                & S2S       & 26.6              & 911.1            & 616.6            & \textbf{250.0}    & \textbf{157.3}   & 392.3          \\
			& S2P       & 25.8              & \textbf{268.3}   & 550.6            & 344.9             & 389.1            & 315.8          \\
			& S2SS      & 25.2              & 741.1            & 1051.9           & 441.2             & 222.9            & 496.5          \\
			& VAE-NILM  & \textbf{21.5}     & \textbf{293.6}   & \textbf{427.2}   & \textbf{271.7}    & \textbf{178.8}   & \textbf{238.5} \\
			&           &                   &                  &                  &                   &                  &                \\[-2ex]
			EpD    & DAE       & 188               & 93        & 140              & 608               & 169              & 240            \\
			(Wh)   & S2S       & 257               & 172       & \textbf{107}     & 317               & 131              & 197            \\
			& S2P       & 209               & \textbf{78}      & \textbf{106}     & 284               & 187              & 173            \\
			& S2SS      & 281               & 112              & 204              & 283               & 121              & 200            \\
			& VAE-NILM  & \textbf{116}      & \textbf{60}      & \textbf{91}      & \textbf{211}      & \textbf{73}      & \textbf{110}   \\
			&           &                   &                  &                  &                   &                  &                \\[-2ex]
			PR                 & DAE       & 76.9              & 47.6             & 20.8             & 48.6              & 65.5             & 51.9           \\
			(\%)               & S2S       & 70.7              & 6.7              & 43.6             & \textbf{97.2}     & \textbf{94.7}    & 62.6           \\
			& S2P       & 77.6              & 57.9             & \textbf{66.4}    & 91.3              & 49.6             & 68.6           \\
			& S2SS      & 66.4              & \textbf{91.5}    & 35.0             & 70.3              & 70.8             & 66.8           \\
			& VAE-NILM  & \textbf{84.9}     & 83.2             & 49.5             & 92.6              & 84.2             & \textbf{78.9}  \\
			&           &                   &                  &                  &                   &                  &                \\[-2ex]
			RE                 & DAE       & 77.8              & \textbf{98.5}    & 60.8             & 26.1              & 28.5             & 58.3           \\
			(\%)               & S2S       & 75.7              & 79.3             & 60.4             & 25.5              & 19.9             & 52.2           \\
			& S2P       & 78.8              & \textbf{96.9}    & 64.4             & 46.1              & 30.5             & 63.3           \\
			& S2SS      & \textbf{83.7}     & 72.7             & 10.0             & \textbf{86.7}     & \textbf{77.9}    & 66.2           \\
			& VAE-NILM  & \textbf{82.2}     & 91.5             & \textbf{68.1}    & 69.8              & 65.1             & \textbf{75.4}  \\
			&           &                   &                  &                  &                   &                  &                \\[-2ex]
			F1                 & DAE       & 77.3              & 64.2             & 31.0             & 34.0              & 39.7             & 49.2           \\
			(\%)               & S2S       & 73.1              & 12.4             & 50.7             & 40.4              & 32.9             & 41.9           \\
			& S2P       & 78.2              & 72.5             & \textbf{65.3}    & 61.3              & 37.7             & 63.0           \\
			& S2SS      & 74.0              & 81.1             & 15.5             & \textbf{77.7}     & \textbf{74.2}    & 64.5           \\
			& VAE-NILM  & \textbf{83.5}     & \textbf{87.2}    & 57.3             & \textbf{79.6}     & \textbf{73.4}    & \textbf{76.2}  \\
			&	       &	     	       &	     	      &	    	         &	     	         &	    	        &	        	 \\[-2ex]
			\bottomrule[2pt]
		\end{tabular}
	}
	\vspace{0.3cm}
	\label{tab:Result_REFIT}
\end{table}

\subsection{State detection analysis}
Overall, the results of the experiment suggest that the proposed model achieves better detection of target appliance states in the aggregate signal with an F1-Score 11\% higher than the reference methods. The \mbox{VAE-NILM} mainly improves the precision metric, by reducing the number of false positives. We hypothesize that a larger window size helps to capture the context of the other active appliances in the aggregate signal and provides a better prediction of the target appliance states. The proposed model supports the use of larger windows and therefore to benefit from these advantages. As shown in Fig. \ref{fig:Activations} for the fridge and kettle, the \mbox{VAE-NILM} generates less residual power activation than reference methods.

\subsection{Energy estimation analysis}
The \mbox{VAE-NILM} obtains on average a more accurate EpD prediction than the reference methods. However, we notice that for some appliances, another method obtains a lower EpD, while the MAE and F1-Score are better for the \mbox{VAE-NILM}. We explain this discrepancy with the EpD definition. The EpD metric is defined as the absolute value of the difference between the predicted and actual energy consumption per day. The weakness of this metric is that the energy consumption used to calculate the EpD is the sum of the power measurements over a period of time. Thus, some false positive and false negative activations cancel each other out over the period of one day. This causes a lower EpD for a model that has more difficulty detecting the target appliance in the aggregate signal. Regarding this, we note that the average F1-Score is higher by more than 11\% for \mbox{VAE-NILM} than the reference methods.

In light of the results reported, we conclude that the \mbox{VAE-NILM} model outperforms state-of-the-art methods for the state detection of appliances with a higher F1-Score \ctony{on both datasets}. Finally, the proposed model yields a more accurate signal power reconstruction than the reference methods, which reduces the overall MAE and more specifically the $\text{MAE}_{ON}$.

\section{Conclusion}
\label{sec:Conclusion}
In this paper, we propose an approach based on the VAE framework for the energy disaggregation task. The regularized latent space of the VAE facilitates the encoding of the relevant features essential for an accurate reconstruction of the target appliance power signal. Furthermore, batch and instance normalization implemented in the \mbox{IBN-Net} helps to stabilize and improve the learning process of the proposed approach. Skip connections between the encoder and decoder layers allow the transfer of the extracted features and contribute to enhancing the reconstruction capability of the decoder. In addition, the proposed model can generate realistic appliance activation by varying the latent variable. Thus, the model is able to create synthetic activations to enhance the training data of energy disaggregation approaches. The proposed model was compared to state-of-the-art NILM approaches on the \mbox{UK-DALE} and REFIT datasets and yielded competitive results. \mbox{VAE-NILM} showed improvements for the detection of the target appliance in the aggregate signal with an average increase of 11\% for the F1-Score on both datasets. Moreover, the proposed model demonstrated a more accurate reconstruction capability, especially for multi-state appliances, \ctony{with an average reduction of 130 and \mbox{77 W} for the $\text{MAE}_{ON}$ on the \mbox{UK-DALE} and REFIT datasets respectively.}

In a future work, we will investigate the integration of multi-task learning (MTL) techniques into this model. MTL would improve both the detection of appliance activations by reducing false positives and false negatives, and the energy disaggregation performance. Furthermore, we expect MTL would increase the generalization capability of the model and yield a better reconstruction for appliances in different houses.

\bibliographystyle{unsrtnat}
\bibliography{biblio}  






\end{document}